\documentclass[twoside]{article}
\pdfoutput=1 
\usepackage[accepted]{aistats2022}
\usepackage[round]{natbib}

\usepackage[utf8]{inputenc} 
\usepackage[T1]{fontenc}    
\usepackage{hyperref}       
\usepackage{url}            
\usepackage{booktabs}       
\usepackage{graphicx}
\usepackage{subfig}
\usepackage{csquotes}
\usepackage{amsmath}
\usepackage{amsfonts}       
\usepackage{nicefrac}       
\usepackage{microtype}      
\usepackage{arydshln}
\usepackage{xcolor}         

\usepackage[algoruled,vlined]{algorithm2e}
\usepackage[export]{adjustbox}
\usepackage{sidecap}
\usepackage{wrapfig}
\usepackage{gensymb}

\renewcommand{\mathbf}{\boldsymbol}
\usepackage{enumerate}
\renewcommand{\mathbf}{\boldsymbol} 
\newcommand{\Reals}{\mathbb{R}}
\newcounter{mycounter}  
\newenvironment{noindlist}
 {\begin{list}{\arabic{mycounter}.~~}{\usecounter{mycounter} \labelsep=0em \labelwidth=0em \leftmargin=0.5em \rightmargin=0.5em \itemindent=0em}}
 {\end{list}}

\begin{document}

%

%
\runningauthor{Schwöbel, Jørgensen, Ober, and van der Wilk}

\twocolumn[

\aistatstitle{Last Layer Marginal Likelihood for Invariance Learning}

\aistatsauthor{Pola Schwöbel \And Martin Jørgensen}

\aistatsaddress{\url{posc@dtu.dk}, DTU \And  University of Oxford} 

\aistatsauthor{Sebastian W. Ober \And Mark van der Wilk }

\aistatsaddress{University of Cambridge \And Imperial College London}]

\begin{abstract}
Data augmentation is often used to incorporate inductive biases into models. Traditionally, these are hand-crafted and tuned with cross validation. The Bayesian paradigm for model selection provides a path towards end-to-end learning of invariances using only the training data, by optimising the marginal likelihood.
Computing the marginal likelihood is hard for neural networks, but success with tractable approaches that compute the marginal likelihood for the last layer only raises the question of whether this convenient approach might be employed for learning invariances. We show partial success on standard benchmarks, in the low-data regime and on a medical imaging dataset by designing a custom optimisation routine. Introducing a new lower bound to the marginal likelihood allows us to perform inference for a larger class of likelihood functions than before. On the other hand, we demonstrate failure modes on the CIFAR10 dataset, where the last layer approximation is not sufficient due to the increased complexity of our neural network. Our results indicate that once more sophisticated approximations become available the marginal likelihood is a promising approach for invariance learning in neural networks.
\end{abstract}

\begin{figure}[t]
 \centering
  \captionsetup[subfigure]{justification=centering}
 \subfloat[M1:\\
            Train MSE: $\mathbf{0.047}$ \\
            Test MSE $0.572$ \\
            Marg. Likelihood $-13.16$ ]{
  \includegraphics[width=0.49\hsize]{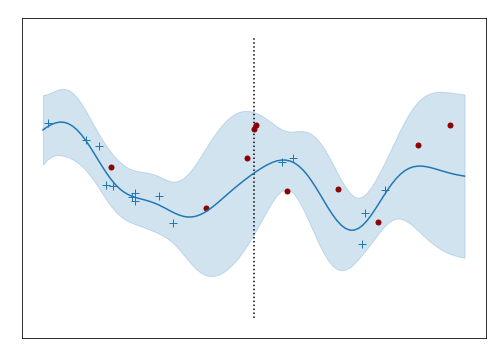}}
  \subfloat[M2: \\
            Train MSE $0.100$ \\ 
            Test MSE $\mathbf{0.235}$ \\
            Marg. Likelihood $\mathbf{-12.2}$]{
   \includegraphics[width=0.49\hsize]{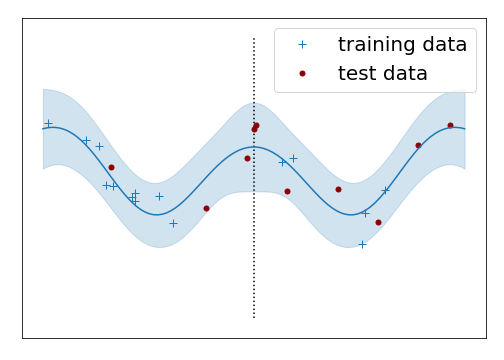}}
 \caption{A non-invariant model M1 and its sign invariant (i.e.\ symmetric around $x=0$) counterpart M2.  The non-invariant M1 has a better train MSE, but the invariant M2 has a better test MSE. The log marginal likelihood correctly identifies M2 as better.} \label{fig:intro}
 \end{figure}

\section{INTRODUCTION} \label{sec:introduction}
Human learners generalise from example to category with seemingly little effort. Machine learning models aim to make accurate predictions on unseen data points based on finitely many examples.

 This generalisation is enabled by \textit{inductive biases}. In \textit{Steps toward Artificial Intelligence} Marvin \citet{minsky1961steps} highlights the importance of invariance as an inductive bias:
`\textit{One of the prime requirements of a good property is that it be invariant under the commonly encountered equivalence transformations. Thus for visual Pattern-Recognition we would usually want the object identification to be independent of uniform changes in size and position.}'
In modern machine learning pipelines invariances are achieved through data augmentation. If we, for example, would like our neural network to be invariant with respect to rotation, we simply present it with rotated versions of the input data. 
Data augmentation schemes are almost always hand-crafted, based on assumptions and expert knowledge about the data, or found by cross-validation. We aim to learn invariances with backpropagation, to reduce the human intervention in the design of ML algorithms.

Learning invariances through gradients requires a suitable loss function. Standard losses like negative log-likelihood or mean squared error solely measure how tightly we fit the training data. 
Good inductive biases (e.g.~convolutions) constrain the expressiveness of a model, and therefore do not improve the fit on the training data. Thus, they can not be learned by minimising the training loss alone. 

In Bayesian inference, this problem is known as \emph{model selection}, and is commonly solved by using a different training objective: the marginal likelihood. For a model of data $y$, parametrised by weights $w$ and hyperparameters $\theta$ it is given by
\begin{equation}\label{eq:marglike}
p(y | \theta) = \int p(y|w) p(w|\theta) \text{d}w\,.
\end{equation}
As opposed to standard training losses, it correlates with generalisation, and thus provides a general way to select an inductive bias, independent of parameterisation
\citep{williams2006gaussian, rasmussen2001occam, mackay2002information}. \Citet{van2018learning} demonstrated that invariances can be learned by straightforward backpropagation using the marginal likelihood in Gaussian process (GP) models, where the marginal likelihood can be accurately approximated. Fig.~\ref{fig:intro} shows an invariant and a non-invariant GP; the invariant model has higher marginal likelihood as well as lower test mean squared error. Thus, the marginal likelihood correctly identifies invariance as a useful inductive bias. 

Current GP models often lack predictive performance compared to their highly expressive neural network counterparts, hence applying this elegant principle to neural networks is attractive. The challenge is, however, that finding accurate and differentiable marginal likelihood approximations for neural networks is still an open problem. In this work we investigate a convenient short-cut: computing Bayesian quantities only in the last layer. This avoids difficulties of the marginal likelihood in the full network, and has already been shown helpful \citep{wilson2016deep,wilson2016stochastic}. Given the possible impact of invariance learning with the convenience of the last-layer approximation, it is important to investigate its potential. Our results provide a nuanced picture of this approach: there are situations where the last-layer approximation is sufficient, but others where it is not. 

To provide these results, we
\begin{enumerate}
    \item construct a \textit{deep neural network with a Bayesian last layer that incorporates invariance}, based on invariant GPs \citep{van2018learning} and deep kernel learning \citep{wilson2016stochastic},
    \item overcome problems with the training implied by a straightforward combination of \Citet{van2018learning} and \citet{wilson2016stochastic} via a \textit{new optimisation scheme}, and \textit{a new variational bound} that allows for non-Gaussian likelihoods,
    \item \textit{investigate failure modes} on more complex model architectures to show limitations of using the last-layer approximation for invariance learning.
\end{enumerate}

\section{RELATED WORK}
\textbf{Bayesian Deep Learning} 
aims to provide principled uncertainty quantification for deep models. Exact computation for Bayesian deep models is intractable, so different approximations have been suggested. Variational strategies \citep[e.g.][]{blundell2015weight} maximise the evidence lower bound (ELBO) to the marginal likelihood, thereby minimising the gap between approximate and true posteriors. To remain computationally feasible, approximations for Bayesian neural networks are often crude, and while weight posteriors are useful in practice, the marginal likelihood estimates are typically too imprecise for hyperparameter estimation \citep{blundell2015weight, turner2011problems}.
Hyperparameter estimation in deep GPs has achieved more success \citep{damianou2013deep, dutordoir2020bayesian}, but training deep GPs can be challenging.
Some very recent works have shown initial promise in using the marginal likelihood for hyperparameter selection in Bayesian neural networks
\citep{ober2020global, immer2021scalable, dutordoir2021deep}. Instead of a Bayesian treatment of all weights using rough approximations, we follow a deep kernel learning approach, i.e.\ computing the marginal likelihood for the last layer only.

\textbf{Deep Kernel Learning} \citep[DKL;][]{hinton2007using, calandra2016manifold, bradshaw2017adversarial} replaces the last layer of a neural network with a GP, where marginal likelihood estimation is accurate \citep{burt2020gpviconv}. \citet{wilson2016deep, wilson2016stochastic} had significant success achieving improved uncertainty estimates. Their results indicate that such a neural network-GP hybrid is promising for invariance learning.
\citet{ober2021promises} identify difficulties with overfitting in DKL models, but also show mechanisms by which such overfitting is mitigated. We find similar issues and adapt the standard DKL training procedure to avoid them when learning invariance hyperparameters. We will discuss these issues in more depth as we describe our training procedure in Sec.~\ref{sec:training_scheme}. 

\textbf{Data Augmentation} is used to incorporate invariances into deep learning models. Where good invariance assumptions are available a priori (e.g. for natural images) this improves generalisation performance and is ubiquitous in deep learning pipelines. Instead of relying on assumptions and hand-crafting, recent approaches \textit{learn} data augmentation schemes. \Citet{cubuk2019autoaugment, cubuk2020randaugment} and \citet{ho2019population} train on the validation data, and use reinforcement learning and evolutionary search respectively to find parameters. \citet{zhou2020meta, lorraine2020optimizing} compute losses on validation sets for learning invariance parameters, and estimate gradients w.r.t.\ them in outer loops. Similar to our work, \citet{benton2020learning} learn data augmentations on training data end-to-end, by adding a regularisation term to the negative log-likelihood loss that encourages invariance. They argue that tuning this regularisation term via cross-validation can be avoided, since the loss function is relatively flat. Yet, the method relies on explicit regularisation, and thus on an understanding of the parameters in question. Our method is based on a Bayesian view of data augmentation as incorporating an invariance on the functions in the prior distribution \citep{van2018learning,nabarro2021data}. This allows the marginal likelihood to be used as an objective for learning invariances. This has many advantages, such as allowing backpropagation from training data, automatic and principled regularisation, and parameterisation independence (see Sec.~\ref{sec:training_scheme}). This makes the marginal likelihood objective a promising avenue for future work, which may want to incorporate invariances whose parameterisations are non-interpretable.

\section{BACKGROUND} \label{sec:background}
\subsection{Variational Gaussian processes}
A Gaussian process (GP) \citep{williams2006gaussian} is a distribution on functions with the property that any vector of function values
$
    \mathbf{f}=\left(f(x_1),\ldots,f(x_N)\right)
$
is Gaussian distributed. We assume zero mean functions and real valued vector inputs.

Inference in GP models with general likelihoods and big datasets can be done with variational approximations  \citep{pmlr-v5-titsias09a,hensman2014scalable}. The approximate posterior is constructed by conditioning the prior on $M$ \emph{inducing variables} $\mathbf u \in \mathbb R^M$, and specifying their marginal distribution with $q(\mathbf{u})=\mathcal{N}(\mathbf{m},\mathbf{S})$ (for overviews see \citealt{bui2017unifying,vdw2020framework}). This results in a variational predictive distribution:
\begin{align}
    q(f(x^*))= \mathcal{N}\big(\mathbf{\alpha}(x^*)^\top\mathbf{m}, \label{eq:q} \\
    k(x^*,x^*) - & \mathbf{\alpha}(x^*)^\top\left(\mathbf{K}_{\mathbf{zz}}-\mathbf{S}\right)\mathbf{\alpha}(x^*)\big), \nonumber
\end{align}
where $\mathbf{z}\!\in\!\Reals^{M\times d}$ are inducing \emph{inputs}, $\mathbf{K}_{\mathbf{zz}}$ is the matrix with entries $k(z_i,z_j)$, $\mathbf{\alpha}(x^*)\!=\!\mathbf{K}_{\mathbf{zz}}^{-1}k(\mathbf{z},x^*)$, and $k$ is the chosen covariance function.

Variational inference (VI) selects an approximation by minimising the KL divergence of the approximation to the true posterior with respect to the variational parameters $\mathbf z, \mathbf m, \mathbf S$. This is done by maximising a lower bound to the marginal likelihood (the ``evidence''), which has the KL divergence as its gap \citep{matthews2016}. The resulting evidence lower bound (ELBO) is
\begin{align}
\log p(y) \geq \mathcal{L} = \sum_{n=1}^N &\mathbb{E}_{q(f(x_n))} \left[\log p(y_n|f(x_n))\right] \nonumber \\
& - \text{KL}[q(\mathbf{u}) || p(\mathbf{u}) ].
\end{align}

In exact GPs, (kernel) hyperparameters are found by maximising the log marginal likelihood $\log p(y)$ \citep{williams2006gaussian}. For our models of interest, the exact marginal likelihood is intractable. We use the ELBO as a surrogate. This results in an approximate inference procedure that maximises the ELBO with respect to both the variational parameters and the hyperparameters. Optimising the variational parameters improves the quality of the posterior approximation, and tightens the bound to the marginal likelihood. Optimising the hyperparameters hopefully improves the model, but the slack in the ELBO can lead to worse hyperparameter selection \citep{turner2011problems}. 

\subsection{Invariant Gaussian Processes} \label{sec:invariant_gps}
A function $f:\mathcal{X}\!\rightarrow\!\mathcal{Y}$ is \textit{invariant} to a transformation $t:\mathcal{X}\!\to\!\mathcal{X}$ if $f(x)\!=\!f(t(x))$, $\forall x \in \mathcal{X}$, and $\forall t \in \mathcal{T}$.
I.e., an invariant function will have the same output for a certain range of transformed inputs known as the \emph{orbit}. A straightforward way to construct invariant functions is to simply average a function over the orbit \citep{kondor2008thesis,ginsbourger2012argumentwise,ginsbourger2013kernels}. We consider a similar construction where we average a function over a data augmentation distribution, which results in an approximately invariant function where $f(x)\!\approx\!f(t(x))$ \citep{van2018learning,dao2019kernel}. Augmented data samples $x_a$ are obtained by applying random transformations $t$ to an input, $x_a = t(x)$, leading to the distribution $p(x_a|x)$.

That is, an approximately invariant function $f$ can be constructed from any non-invariant $g$ as
\begin{equation} \label{eq:invariant_function_integral}
    f(x)\!=\! \sum_{t \in \mathcal{T}} g(t(x)), \text{ or } f(x)\!=\!  \int  g(x_a) p(x_a | x) \text{d}x_a.
\end{equation}
\Citet{van2018learning} exploit this construction to build a GP with continuously adjustable invariances. They place a GP prior on $g\sim\mathcal{GP}\left(0,k_g(\cdot,\cdot)\right)$, and since Gaussians are closed under summations, $f$ is a GP too. By construction $f$ is invariant to the augmentation distribution $p(x_a|\cdot)$ and its kernel is given by\looseness=-1 
\begin{equation} \label{eq:invariant_kernel}
    k_f(x,x')\!=\! \iint  k_g(x_a,x'_a) p(x_a|x) p(x'_a|x') \text{d}x_a\text{d}x'_a.
\end{equation}
Non-trivial $p(x_a|x)$ densities present a problem for standard VI, as the kernel evaluations in eq.~\ref{eq:q} become intractable.  This is solved by making the inducing variables observations of $g$ rather than the usual $f$. This ensures that $\mathbf{K}_{\mathbf{zz}}$ is tractable, as it only requires evaluations of $k_g$, which makes the KL divergence tractable. When the likelihood is Gaussian, it additionally provides a way to tackle the expected log likelihood:
\begin{align}
\mathbb{E}_{q(f(x))}\log \mathcal{N}(y; f(x),\sigma^2) = \text{const} \!-\! \frac{(y_n\!-\!\mu)^2 \!+\! \tau}{2\sigma^2}
\end{align}
where $\mu,\tau$ are the mean and variance in \eqref{eq:q}. Only unbiased estimates of $\mu$, $\mu^2$ and $\tau$ are needed for an unbiased estimate of the ELBO. These can be obtained from simple Monte Carlo estimates of $k_f$ \eqref{eq:invariant_kernel}, and $k(\mathbf z, x)$.\footnote{We obtain $k(\mathbf z, x) \!\!=\!\! \int k_g(\mathbf z, x_a)p(x_a|x)\mathrm dx_a$ from the interdomain trick, which can be estimated with Monte Carlo. See \Citet{van2018learning} for details.}

\begin{figure*}[t]
 \centering
  \includegraphics[width=0.75\hsize]{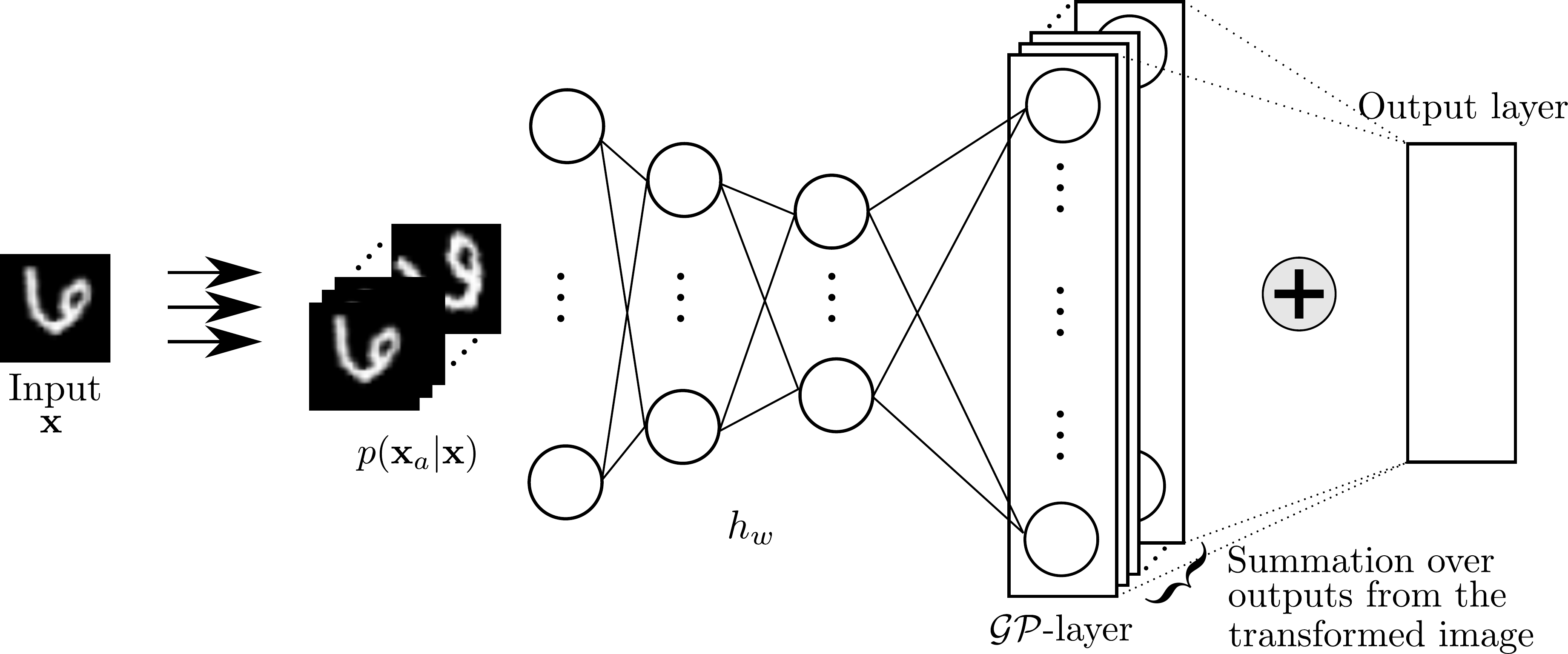}
 \caption{A visualisation of the model pipeline. For any input $x$, we can sample from the orbit distribution $p(x_a|x, \phi)$; each of these sample gets passed through a neural network parametrised by $w$. The last layer is a of the net is a GP, on which we can sum across sample outputs to create an invariant function.}  \label{fig:model}
 \end{figure*}
\subsection{Parameterising learnable invariances} \label{sec:image_trafos}
The invariance of the GP in \eqref{eq:invariant_kernel} is learned by adjusting the augmentation distribution. We parameterise the distribution and treat its parameters as kernel hyperparameters. We learn these by maximising the ELBO. As done in similar work \citep{benton2020learning, van2018learning}, we consider affine transformations. Our affine transformations are controlled by $\phi = (\alpha, s^x, s^y, p^x, p^y, t^x, t^y)$, which describes rotation, scale, shearing and horizontal and vertical translation. We parameterise a family of augmentation distributions by specifying uniform ranges with $\phi_\text{min}, \phi_\text{max} \in \Reals^7$ that are to be applied to the input image. Different ranges that are learned on $\phi_{\text{min}}, \phi_\text{max}$ correspond to different invariances in $f(\cdot)$. For example, learning $\phi_{\text{min/max}} = (\pm \pi, 0, 0, 0, 0, 0, 0)$ corresponds to full rotational invariance (sampling any angle between $-\pi$ and $\pi$) but no scaling, shearing or translations.

We sample from the resulting $p(x_a|x, \phi_\text{max}, \phi_\text{min})$ (we will write $p(x_a|x, \phi)$ for brevity) by \textbf{1)} sampling the parameters for a transformation from a uniform distribution, \textbf{2)} generating a transformed coordinate grid, and \textbf{3)} interpolating\footnote{Image transformation code from \url{github.com/kevinzakka/spatial-transformer-network}} the image $x$:
\begin{align}
    x_a = t_\nu(x)\,, && \nu \sim U(-\phi_{\text{min}}, \phi_\text{max}) \,.
\end{align}
Since transforming $t_\nu(x)$ is differentiable, this procedure is reparameterisable w.r.t.\ $\phi_\text{max}, \phi_\text{min}$ via $\nu = \phi_\text{min} + (\phi_\text{max} - \phi_\text{min})\varepsilon, \ \varepsilon \sim U(0, 1)$.
Straightforward automatic differentiation of the unbiased ELBO estimator described in the previous section provides the required gradients. 

In summary, we learn $\phi_{\text{min/max}}$ by maximising the ELBO, so the transformations and their magnitudes are learned based on the specific training set. Different invariances will be learned for different training data. The next sections show how these principles have potential even in neural network models, beyond the single layer GPs of \Citet{van2018learning}.

\begin{algorithm}
\SetAlgoLined
\begin{noindlist}
    \item Draw $S$ samples from the augmentation distribution $x^i_a \sim p(x_a | x, \phi),  \ i=1 ... S$.
    \item Pass the $x_a^i$ through the neural net $h_w$. 
    \item Map extracted features using the non-inv. $g$.
    \item Aggregate samples to obtain inv. $f(x)$ by 
    \begin{enumerate}[(i) ]
    \item using the unbiased estimators from Sec.~\ref{sec:invariant_gps} in the Gaussian case, or,
    \item averaging predictions $g(h_w(x_a^i)), \ i=1, ..., S$  directly in the Softmax case, see \eqref{eq:fhat}.
    \end{enumerate}
\end{noindlist}
 \caption{InvDKGP forward pass} \label{InvDKGP_algo}
\end{algorithm}

\section{MODEL}\label{sec:model}
As discussed in Sec.~\ref{sec:introduction}, we aim to learn neural network (NN) invariances through backpropagation, in the same way as is possible for single-layer GPs.
Since finding high-quality approximations to the marginal likelihood of a NN is an ongoing research problem, we investigate whether a simpler \emph{deep kernel} approach  is sufficient. This uses a GP as the last layer of a NN, and takes advantage of accurate marginal likelihood approximations for the GP last layer. Success with such a simple method would significantly help automatic adaptation of data augmentation in neural network models. We hypothesise that the last layer approximation is sufficient, since data augmentation influences predictions only in the last layer (in the sense that one can construct an invariant function $f$ from an arbitrary non-invariant $g$ by summing in the last layer, eq. \ref{eq:invariant_function_integral}). See Fig.~\ref{fig:model} for a graphical representation and Algorithm \ref{InvDKGP_algo} for forward pass computations.

\textbf{Deep Kernels} take advantage of covariance functions being closed under transformations of their input. That is, if $k_g(\cdot,\cdot)$ is a covariance function on $\mathbb{R}^D\times \mathbb{R}^D$, then $k_g(h_w(\cdot), h_w(\cdot))$ is a covariance function on $\mathbb{R}^d\times \mathbb{R}^d$ for mappings $h_w:\mathbb{R}^d\rightarrow\mathbb{R}^D$.
In our case, $h_w$ is a NN parametrised by weights $w$, and hence $w$ are viewed as hyperparameters of the kernel. The GP prior becomes \begin{equation}\label{eq:g-prior}
    p(g) = \mathcal{GP}\left(0, k_g(h_w(\cdot), h_w(\cdot))\right).
\end{equation}
The idea is to learn $w$ along with the kernel hyperparameters. Importantly, this model remains a GP and so the inference described in Sec.~\ref{sec:background} applies.

\textbf{Our invariant model} combines the flexibility of a NN $h_w(\cdot)$ with a GP $g$ in the last layer, while ensuring overall invariance using the construction from \eqref{eq:invariant_function_integral}:
\begin{equation} \label{eq:model_def}
    f(x) = \int g(h_w(x_a))p(x_a | x, \phi) \text{d}x_a.
\end{equation}
Thus, combining \eqref{eq:invariant_kernel} and \eqref{eq:model_def}, $f$ is an \emph{invariant} GP with a \emph{deep} kernel given as
\begin{align}
    k_f(x,x')=&\!\!\int\!  k_g\big(h_w(x_a),h_w(x'_a)\big) \nonumber \\
    &\qquad p(x_a|x, \phi) p(x'_a|x', \phi) \text{d}x_a\text{d}x'_a.
\end{align} The model is trained to fit observations $y$ through the likelihood function
$p\left(y|f(x)\right)$, where we assume observations $y_i$ are independent conditioned on the marginals $f(x_i)$.

Initially, we investigate training a model by simply combining the invariant GP training objective for Gaussian likelihoods \citep{van2018learning} with standard deep kernel learning  \citep{wilson2016deep,wilson2016stochastic}. 
However, as we will discuss, several issues prevent these training procedures from working. In following sections we investigate why, provide solutions, and introduce a new ELBO that is suitable for more general likelihoods which improves training behaviour. We refer to our model as the \textit{Invariant Deep Kernel GP (InvDKGP)}. An implementation can be found at \url{https://github.com/polaschwoebel/InvDKGP}.

\section{DESIGNING A TRAINING SCHEME}
 \label{sec:training_scheme}
\begin{figure}
 \centering
  \captionsetup[subfigure]{justification=centering}
 \subfloat{
  \includegraphics[width=\hsize]{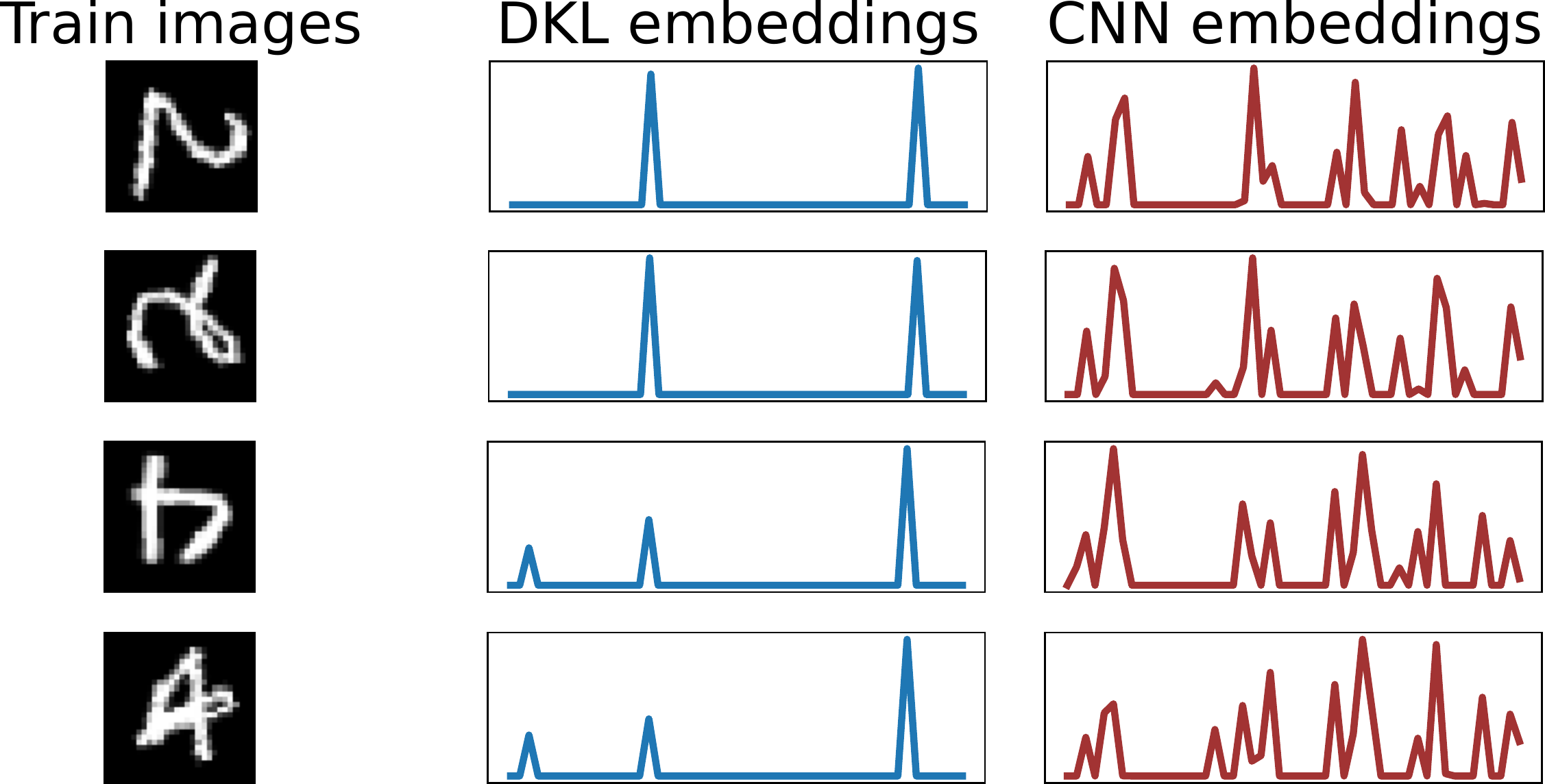}}  
 \caption{Training images with different orientations and their embeddings. Embeddings produced by joint Deep Kernel Learning (DKL, middle column) are similar for all inputs from one class. Little improvement can be gained on the training data by being rotationally invariant. NN embeddings on the right differ depending on input orientation -- signal to learn $p(x_a|x, \phi)$ from.} \label{fig:training_scheme}
 \end{figure}

 The promise of deep kernel learning as presented by \citet{wilson2016deep, wilson2016stochastic} lies in training the NN and GP hyperparameters \textit{jointly}, using the marginal likelihood as for standard GPs.\footnote{Given that this quantity is difficult to approximate, we verify experimentally that we indeed need it and cannot use a simple NN with max-likelihood (see Appendix).} However, prior works have noted shortcomings of this approach \citep{ober2021promises, bradshaw2017adversarial, van2021feature}: the DKL marginal likelihood correctly penalises complexity for the last layer only, while the NN hyperparameters can still overfit. 
 In our setting, i.e.\ when trying to combine deep kernel learning with invariance learning, joint training produces overfit weights which results in simplistic features with little intra-class variation\footnote{This behavior makes sense: The DKL marginal likelihood only penalises complexity in the last layer, (i.e.\ the GP). The simplistic features from Fig. \ref{fig:training_scheme} can be classified by a simple function in the last layer, thus the complexity penalty is small, and the solution has high marg.\ likelihood.}. In particular, all training points from the same class are mapped to very similar activations, independent of orientation. This causes a loss of signal for the invariance parameters (see Fig.~\ref{fig:training_scheme}).
 
 \begin{figure}
  \centering
     \includegraphics[width=0.36\textwidth]{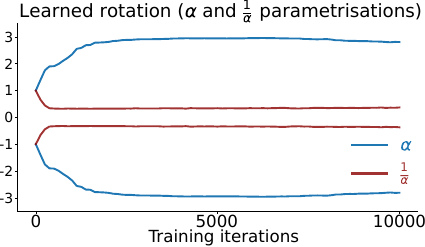}  
 \caption{Learned rot. angles parametrised by $\alpha$ and $\frac{1}{\alpha}$. The $\alpha$-parametrisation, in blue, learns rotational invariance w.r.t. $\pm 2.8$ radians. The  $\frac{1}{\alpha}$-parametrisation (red) learns invariance w.r.t. $\pm \frac{1}{0.37} = \pm 2.7$ radians.}
 \label{fig:angle_inverse_angle}
  \end{figure}
 
 \paragraph{Coordinate ascent training} fixes this problem. We pre-train the NN using negative log-likelihood loss. Then, we replace the fully connected last layer with an invariant GP. The marginal likelihood is a good objective given fixed weights (we obtain a GP on transformed inputs), so we fix the NN weights. However, \textit{some} adaptation of the NN to the transformed inputs is beneficial.  We thus continue training by alternating between updating the NN, and the GP variational parameters and orbit parameters, hereby successfully learning invariances. (See Fig. \ref{fig:MNIST_orbits} and \ref{fig:rotMNIST_orbits}: flat parts of the training curves indicate NN training where all kernel hyperparameters, including invariances, remain fixed. When to toggle between the GP and NN training phase is determined using validation data.)

  \begin{figure}[b]
    \centering
   \includegraphics[width=0.95\hsize]{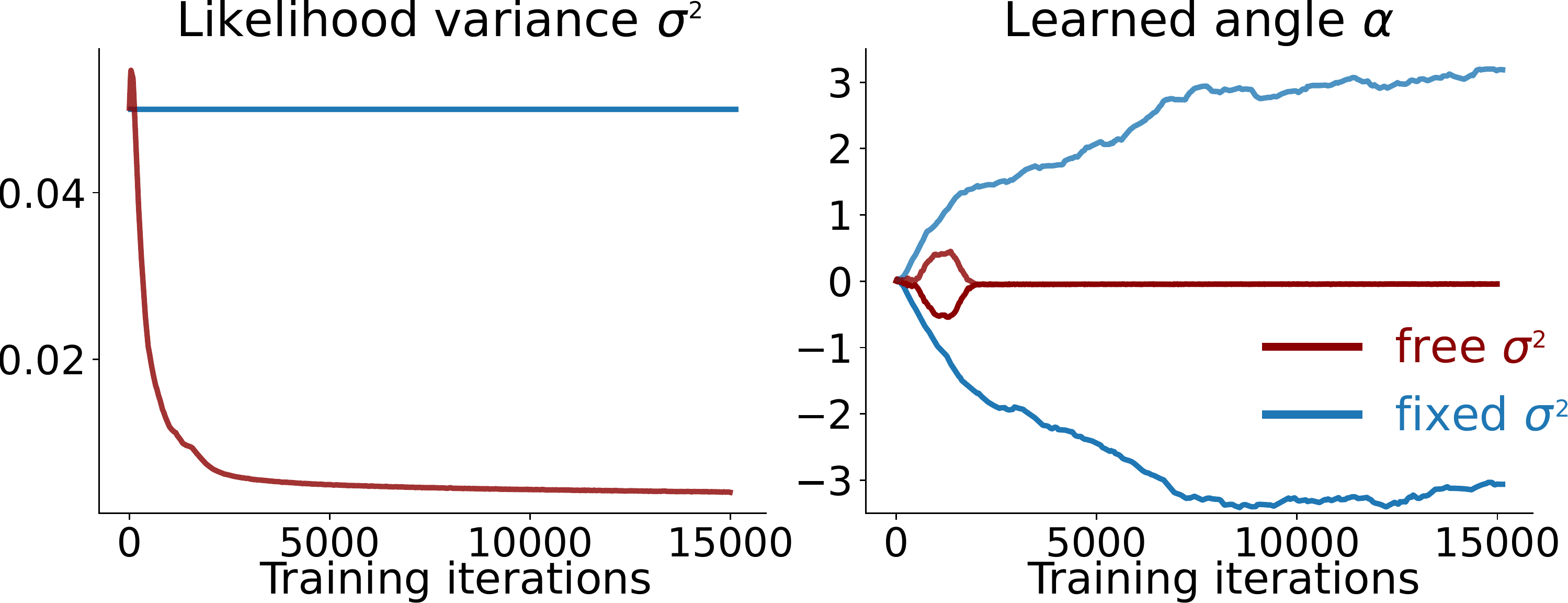}
 \caption{Runs with fixed (red) and non-fixed (blue) kernel and likelihood variance on rotMNIST. The augmentation distribution collapses for non-fixed variances.} \label{fig:fixed_variances}
 \end{figure}

\paragraph{Choosing an invariance parameterisation} is simple with our method. Other invariance learning approaches, e.g. \citet{benton2020learning} and \citet{schwobel2020probabilistic} rely on explicitly regularising augmentation parameters to be large, and thus require interpretability of their parameters. The marginal likelihood objective is \emph{independent of parameterisation}. To illustrate this we compare parameterising the range of angles by the angle in radians $\alpha$ and by its reciprocal $\xi = \frac{1}{\alpha}$. In the rotMNIST example (see Fig~\ref{fig:angle_inverse_angle}) large invariances are needed. This corresponds to large $\alpha$ or small $\xi$ -- our method obtains this in both parameterisations. In contrast, explicitly regularising invariance parameters to be large would fail for $\xi$. We wish to stress that generating the orbit distributions is not restricted to affine image transformation and  parameterisation independence will be more important as more complicated, non-interpretable invariances are considered.

\paragraph{The Gaussian likelihood} is chosen by \Citet{van2018learning} due to its closed-form ELBO. For classification problems, this is a model misspecification. The penalty for not fitting the correct label value becomes large and we can therefore overfit the training data. To alleviate this problem, we fix likelihood and kernel variance (see Fig.~\ref{fig:fixed_variances}). The fixed values were determined by trying out a handful candidates -- this was sufficient to make invariance learning work. To remove this manual tuning, we will derive an ELBO that works with likelihoods like Softmax in Sec~\ref{sec:new_bound}. 

\subsection{MNIST subsets -- the low data regime} \label{sec:MNIST}
\begin{figure}
 \centering
  \includegraphics[width=0.9\hsize]{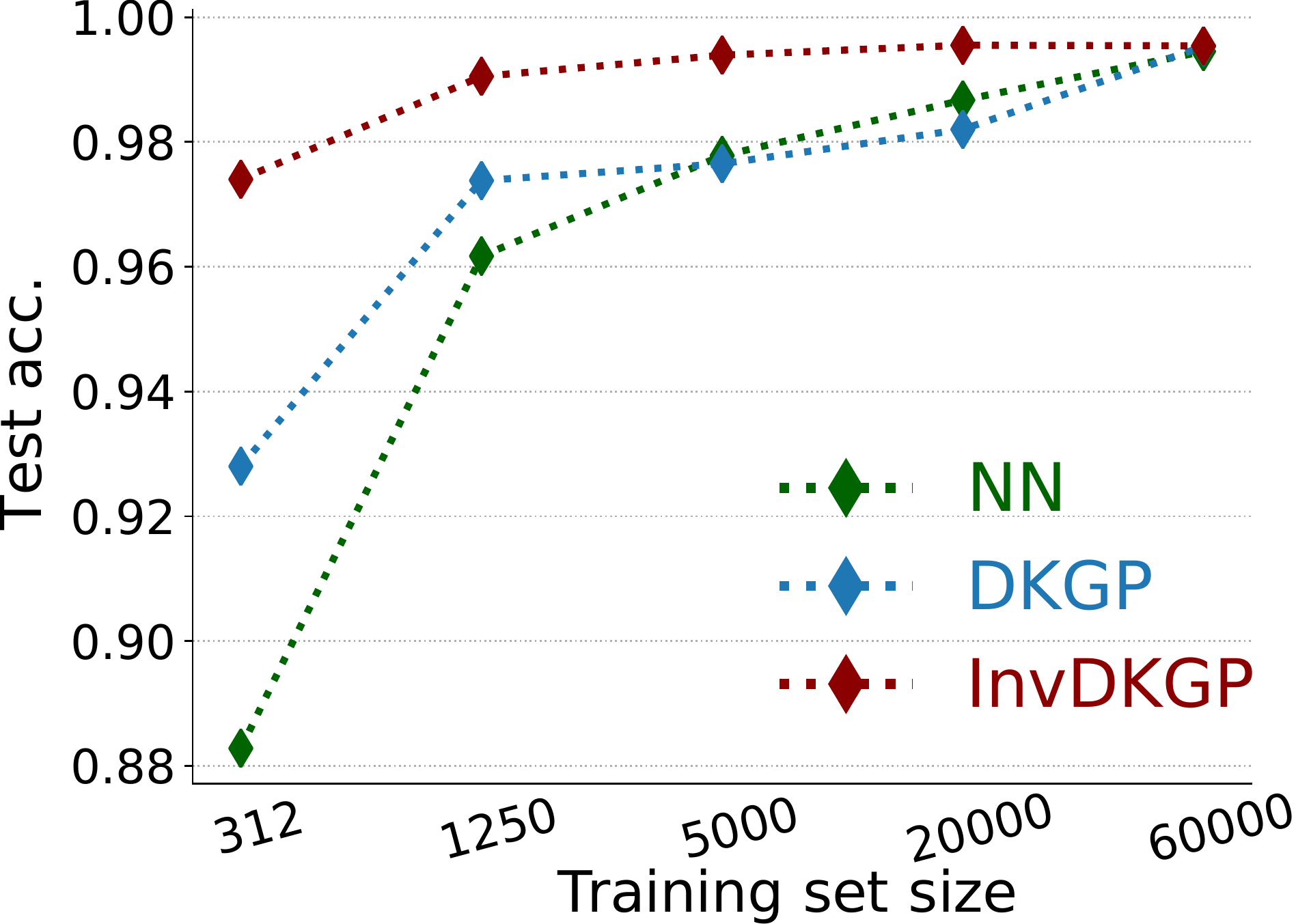}
\caption{Test accuracies against the training set size on MNIST. We see the invariant model (in red) generalises significantly better, especially for small training sets.}
\label{fig:MNIST_subset_accs}
\end{figure}

\begin{figure}[t]
 \centering
 \includegraphics[width=\hsize]{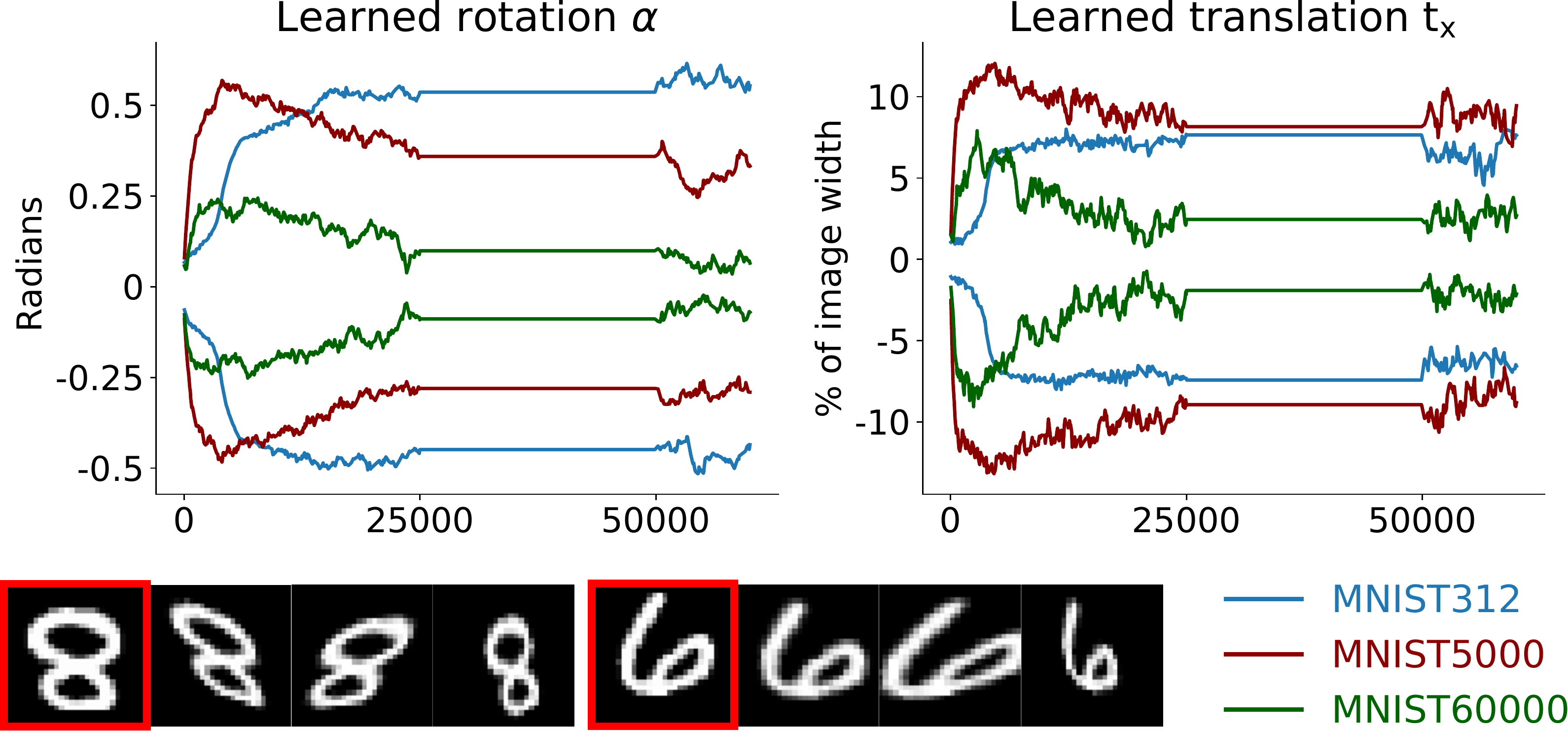}

 \caption{\textit{Top:} Learned invariance parameters (rotation $\alpha$ in radians and x-translation $t_x$) for a small, medium and large training set. We learn larger $\alpha$ for the smaller subsets. Here, data augmentation is more beneficial. \textit{Bottom}: Two training images $x$ (red frames) and samples from $p(x_a | x, \phi)$ (following columns) learned by the InvDKGP on MNIST using only 312 images.} \label{fig:MNIST_orbits}
 \end{figure}

Having developed a successful training scheme we evaluate it on MNIST subsets. The generalisation problem is particularly difficult when training data is scarce. Inductive biases are especially important and usually parameter-rich neural networks rely on heavy data augmentation when applied to smaller datasets. We train on different subsets of MNIST \citep{lecun2010mnist}. InvDKGPs outperform both NNs and non-invariant deep kernel GPs. The margin is larger the smaller the training set --- with only 1250 training examples we can nearly match the performance of a NN trained on full MNIST (Fig.~\ref{fig:MNIST_subset_accs}). We conclude it is possible to learn useful invariances even from small data (see Fig.~\ref{fig:MNIST_orbits}). This data efficiency is desirable since models trained on small datasets benefit crucially from augmentation.

\section{CORRECTING MODEL MISSPECIFICATION}\label{sec:new_bound}

The key observation for inference under the Gaussian likelihood was the \emph{unbiasedness} of the estimators. In this section, we introduce a controlled bias to allow for easy inference in a wide class of likelihoods. In the limit of infinite sampling, the bias disappears and the invariance does not add additional approximation error.

Recall that $f(x)$ constructed in \eqref{eq:invariant_function_integral} is intractable but can be estimated by Monte Carlo sampling
\begin{equation}
    \hat{f}(x):=\frac{1}{S_o}\sum_{i=1}^{S_o} g(x_a^i),
\end{equation}
where $x^i_a\sim p(x_a|x, \phi)$. Notice, 
\begin{equation}
f(x)=\mathbb{E}_{\prod_{i=1}^{S_o} p(x_a^i|x, \phi)}\left[\hat{f}(x)\right]=:\tilde{\mathbb{E}}\left[\hat{f}(x)\right], 
\end{equation}
where $\prod_{i=1}^{S_o} p(x_a^i|x, \phi)$ is the product density over $S_o$ orbit densities. We remark that $f$ is deterministic in $x$ but stochastic in $g$, which is a GP. Thus, we can write
\begin{align}
    \mathbb{E}_{q(f(x))} [\log p(y | & f(x))]=\mathbb{E}_{q(g)}[\log p(y | f(x))]\\
    &=\mathbb{E}_{q(g)}\left[\log p\left(y \big| \tilde{\mathbb{E}}[\hat{f}(x)]\right)\right] \\ 
    & \geq \mathbb{E}_{q(g)}\left[\tilde{\mathbb{E}}\left[\log p\left(y \big| \hat{f}(x)\right)\right]\right]\label{eq:new-bound}.
\end{align}
The inequality is due to Jensen's inequality if the likelihood is \emph{log-concave} in $f$.\footnote{\citet{nabarro2021data} use this same construction in the weight-space of neural networks to find valid posteriors in the presence of data augmentation, although without invariance learning.} This holds for many common likelihoods, e.g. Gaussian and Softmax.

Equality holds above when $\text{Var}(\hat{f}(x))=0$, i.e. the bound becomes tighter as $S_o$ increases \citep[see also][]{burda2015importance}. Hence aggressive sampling recovers accurate VI. The right-hand side of \eqref{eq:new-bound} can now, without additional bias, be estimated by
\begin{equation} \label{eq:fhat}
    \frac{1}{S_g} \sum_{k=1}^{S_g}  \frac{1}{S_{\mathcal{A}}} \sum_{j=1}^{S_{\mathcal{A}}} \log p\left(y \Big| \frac{1}{S_o} \sum_{i=1}^{S_o} g_k(x_a^{ji})\right).
\end{equation}
Since extensive sampling is required to keep the bound above tight, it is important to do this efficiently. From a GP perspective this is handled with little effort by sampling the approximate posteriors $q(g)$ using Matheron's rule \citep{wilson2020efficiently}. Thus, sampling $S_g$ GPs is cheap compared to sampling from the orbit. $S_\mathcal{A}$ denotes the number of $\hat{f}$ samples, this can be fixed to $1$ as long as $S_o$ is large. 

Summarising, we have shown how we can infer through the marginal likelihood, for the wide class of log-concave likelihoods, by maximising the stochastic ELBO:
\begin{align}
    \mathcal L &= \frac{1}{S_g} \sum_{k=1}^{S_g}  \frac{1}{S_{\mathcal{A}}} \sum_{j=1}^{S_{\mathcal{A}}} [\log p(y | \frac{1}{S_o} \sum_{i=1}^{S_o} g_k(h_w(x_a^{ji})))] \nonumber \\
    & \qquad \qquad \qquad-\text{KL}[q(\mathbf{u}) || p(\mathbf{u}) ] \,, \\
    &\text{with } x_a^{ij} \sim p(x_a|x,\phi) \,.
\end{align}

The benefits of our new sample based bound are three-fold: It broadens model specification, avoids hand-picking and fixing the artificial Gaussian likelihood variance, and doubles training speed.

\begin{figure} 
\centering
    \includegraphics[width=\hsize]{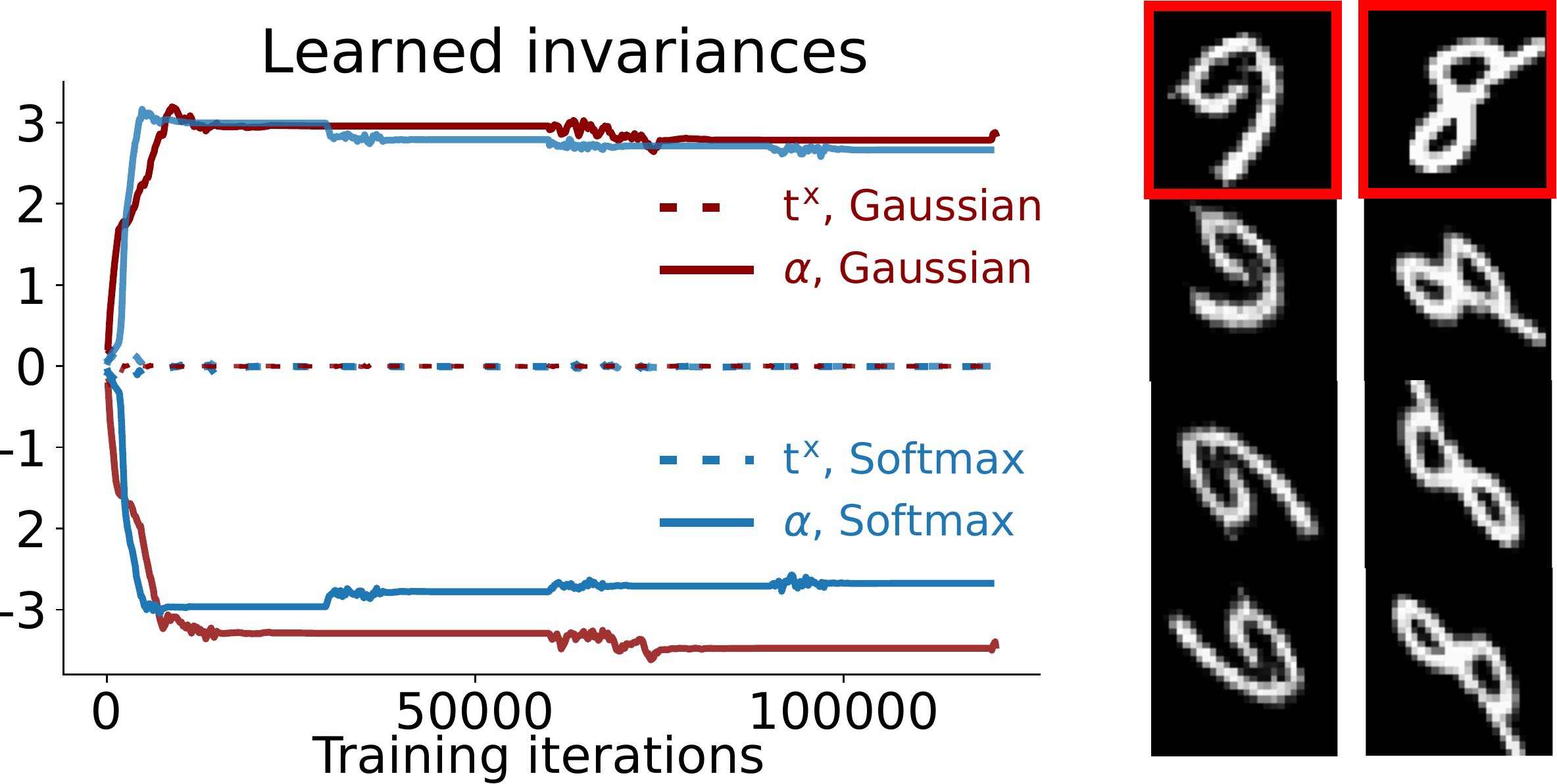}
    \caption{\textit{Left}: Learned invariance parameters (rotation $\alpha$ in radians and x-translation $t_x$) for rotMNIST. Both the Gaussian and the Softmax model learn to be almost fully rotationally invariant (i.e.\ $\alpha_{\text{min/max}} \approx \pm \pi$), and not to be invariant w.r.t.\ translation (i.e.\ $t^x_{\text{min/max}} \approx 0$). Note the different scaling of the $y$-axis to Fig.~\ref{fig:MNIST_orbits}. \textit{Right}: Two training images (red frames) and samples from orbits.}
    \vspace{-1em}
    \label{fig:rotMNIST_orbits}
\end{figure}

\subsection{Rotated MNIST} \label{sec:rotMNIST}
The rotated MNIST dataset\footnote{\url{https://sites.google.com/a/lisa.iro.umontreal.ca/public_static_twiki/variations-on-the-mnist-digits}} 
was generated from the original MNIST dataset by randomly rotating the images of hand-written digits between $0$ and $2\pi$ radians. It consists of a training set of $12.000$ images along with $50.000$ images for testing. We pretrain the neural network from Sec.~\ref{sec:MNIST} on rotated MNIST (Table \ref{table:rotMNIST_results}, M1) and proceed as outlined in Sec.~\ref{sec:training_scheme}. As discussed in Sec.~\ref{sec:training_scheme}, we do not have guarantees that the ELBO acts as a good model selector for the neural network hyperparameters. We thus use a validation set ($3000$ of the $12000$ training points) to find hyperparameters for the NN updates. Once a good training setting is found we re-train on the entire training set (see Appendix for settings). Fig.~\ref{fig:rotMNIST_orbits} shows the learned invariances (we use the full $\phi$ parameterisation but only plot rotation and $x$-translation for brevity). Both Gaussian and Softmax models learn to be rotation-invariant close the full $2\pi$ rotations present in the data. Table \ref{table:rotMNIST_results} contains test accuracies. Deep kernel GPs outperform their shallow counterparts by large margins (differences in test accuracy of $\geq 10$ percent points). The same is true for invariant compared to non-invariant models ($\geq 3$ percent points). While both likelihoods achieve similar test accuracies, we observe a $2.3 \times$ speedup per iteration in training for the sample-based Softmax over the Gaussian model. (Gaussian model: $2.64$ seconds per iteration, Gaussian + sample bound: $1.32$ sec./iter., Softmax + sample bound: $1.13$ sec./iter. All runs are executed on 12 GB Nvidia Titan X/Xp GPUs.)

\begin{table}
\centering
\resizebox{\hsize}{!}{
\begin{tabular}{|l|l|l|c| }  
    \hline
     & \textbf{Model}       & \textbf{Likelihd.}    &\textbf{ Test acc.} \\
    \hline 
    M1 & NN         & Softmax           & 0.9433  \\
    \hline 
    M2 & Non-inv. Shallow GP  & Gaussian          & 0.8357 \\
    M3 & Non-inv. Shallow. GP         & Softmax      &  0.7918 \\
    M4 & Inv. Shallow GP          & Gaussian           & 0.9516 \\
    M5 & Inv. Shallow. GP          & Softmax           & 0.9316  \\
    \hline 
    M6 & Non-inv. Deep Kernel GP & Gaussian      & 0.9387 \\
    M7 & Non-inv. Deep Kernel GP & Softmax       & 0.9351 \\
    M8 & Inv. Deep Kernel GP & Gaussian          & \textbf{0.9896} \\
    M9 & Inv. Deep Kernel GP & Softmax           & 0.9867 \\
    \hline 
\end{tabular}}
\caption{Test accuracies on rotated MNIST. Invariant models outperform non-invariant counterpart. So do deep kernels contra shallow ones. The invariant deep kernel GPs perform best, outperforming state-of-the-art of 0.989 for learned invariance  \citep{benton2020learning}.} \label{table:rotMNIST_results}
\end{table}

\subsection{PatchCamelyon} \label{sec:PCAM}
\begin{figure}
 \centering
 \begin{minipage}{\hsize}
  \subfloat{\includegraphics[width=0.49\hsize, valign=c]{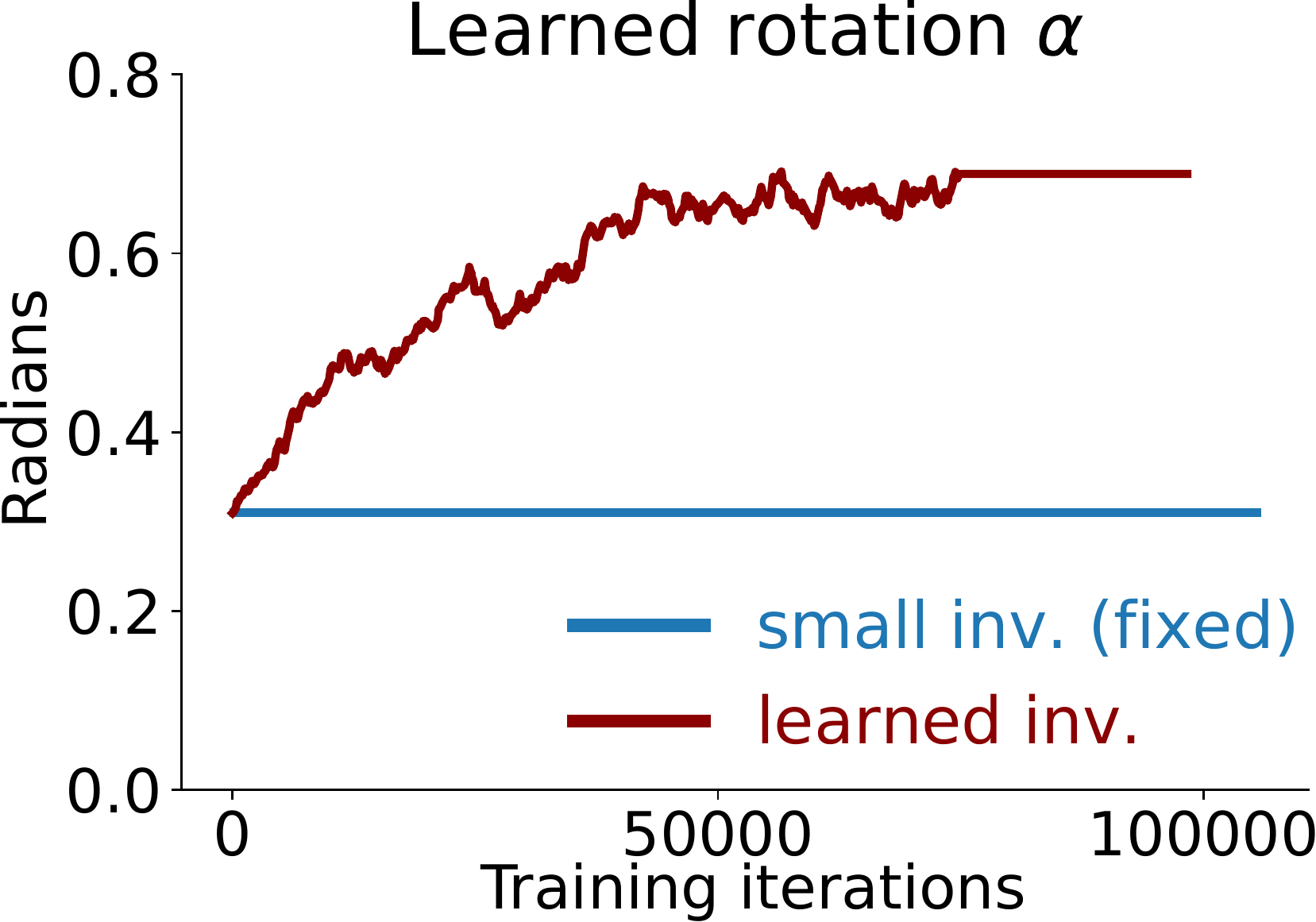}} \hfill
    \subfloat{
  \includegraphics[width=0.49\hsize, valign=c]{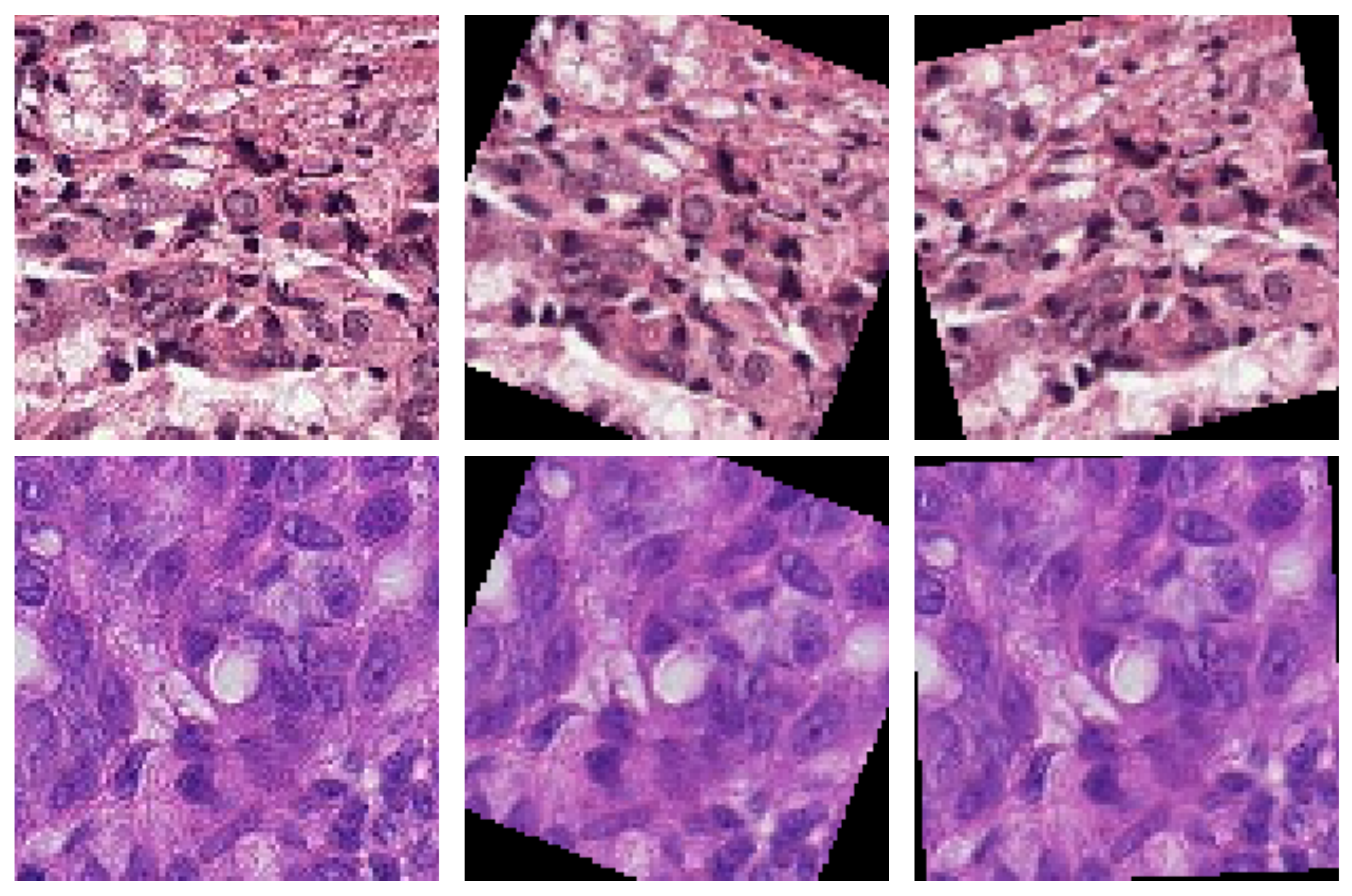}}
  \caption{\textit{Left:} Learned rotation on PCam. \textit{Right:} PCam orbit samples. Augmented images look smoother due to interpolation, thus we preprocess the dataset with small rotations when learning invariances.}
  \vspace{1em}
  \label{fig:PCAM}
  \end{minipage}
  \hfill
  \begin{minipage}{\hsize}
 \centering

 \resizebox{0.8\hsize}{!}{
 \begin{tabular}{|l|l| }  
    \hline
     \textbf{Model}      &\textbf{ Test acc.} \\ \hline
     NN    & 0.7905                        \\
     Deep Kernel GP + no inv. &   0.8018    \\ \hline
     NN + small inv.  &       0.7420       \\ 
     Deep Kernel GP + small inv. &  0.8115    \\ 
     Deep Kernel GP + learned inv.  &    \textbf{0.8171 }  \\ 
    \hline 
\end{tabular}}
\captionof{table}{PCam results. InvDKGP performs best.}
\label{table:PCAM}
\end{minipage}\vspace{-1em}
\end{figure}

The PatchCamelyon (PCam, CC0 License, \cite{Veeling2018-qh}) dataset consists of histopathology scans of lymph nodes measuring $96 \times 96 \times 3$ pixels. Labels indicate whether the centre patch contains tumor pixels. \Citet{Veeling2018-qh} improve test performance from $0.876$ to $0.898$ by using a NN which is invariant to (hard-coded) 90\textdegree\ rotations of the input. 
Such discrete, non-differentiable augmentations are not compatible with our backprop-based method, so we instead use continuously sampled rotations (a special case of the transformations described in Sec.~\ref{sec:image_trafos} with $\phi=\alpha$ and $\alpha_{min}$ = -$\alpha_{max}$). This, contrary to \Citet{Veeling2018-qh}'s approach, introduces the need for padding and interpolation (see Fig. \ref{fig:PCAM}, left), effectively changing the data distribution. We thus apply small rotations as a pre-processing step ($\pi/10$ radians, `small inv.' in Table~\ref{table:PCAM}). This lowers performance for a NN alone, i.e. when pre-training. The invariant models counterbalance this performance drop, and the learned invariances produce the best results in our experiments; however, they remain subpar to \citet{Veeling2018-qh}. This is due to the limitation to differentiable transformations, as well as our simpler NN (see Appendix). We highlight that our task is fundamentally different: instead of hard-coding invariances we \textit{learn} those during optimisation.

\section{EXPLORING LIMITATIONS} \label{sec:conclusion_limitations}

Rotated MNIST and PCAM are relatively simple datasets that can be modelled using small NNs. To investigate whether our approach can be used on more complex datasets, we turn to CIFAR-10 \citep{krizhevsky2009learning}, which is usually trained with larger models and data augmentation. Unfortunately, we found that we were unable to learn invariances for CIFAR-10.

\begin{figure}[t]
 \centering
  \includegraphics[width=0.8\hsize]{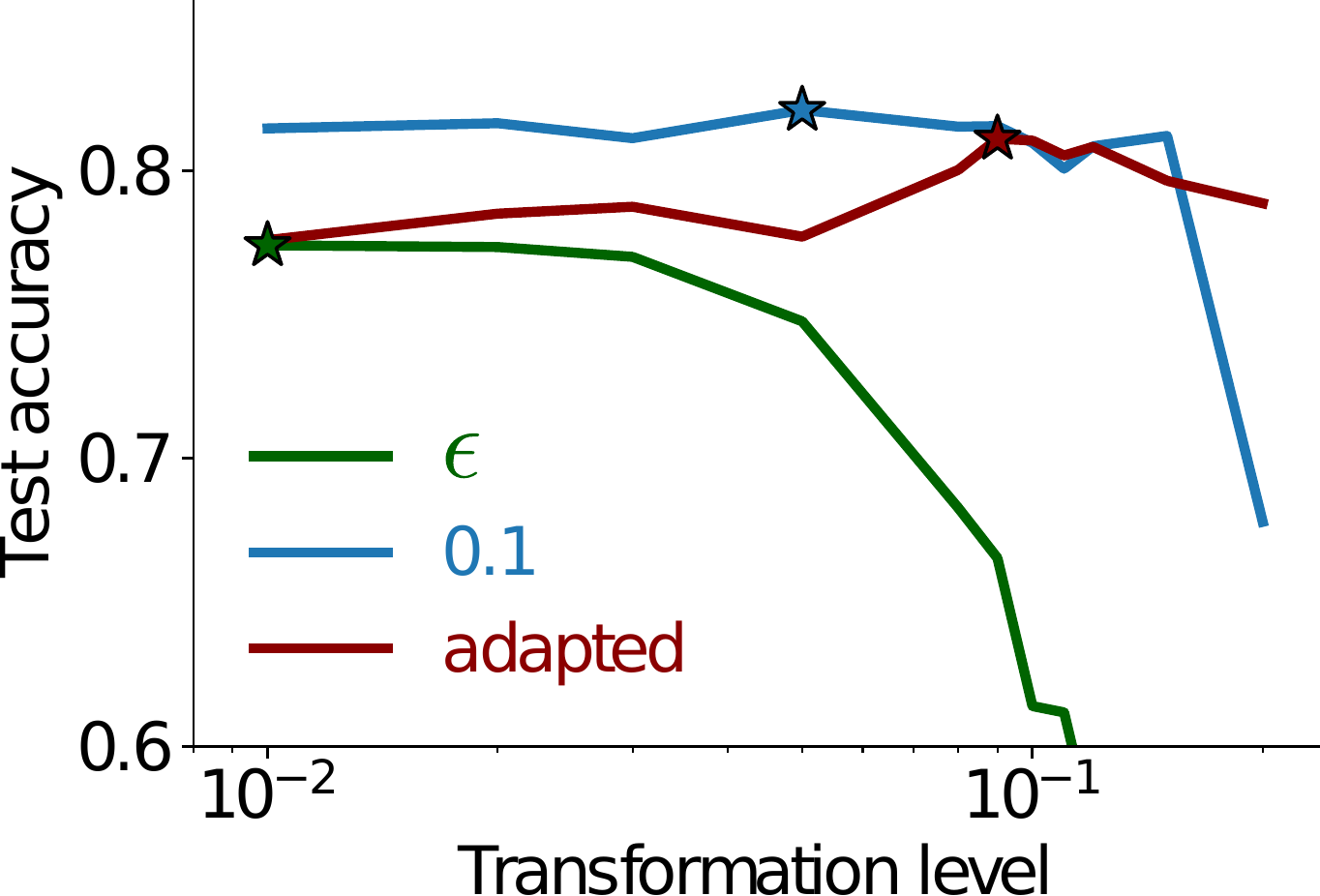}
\caption{Test accuracies on CIFAR-10 for different transformation levels $\nu$ with pretraining at negligible ``$\epsilon$'', 0.1, and adapted levels. The maxima for each curve are marked with a star, and occur at test accuracies of 77.4\% for ``$\epsilon$'', 82.1\% for 0.1, and 81.1\% for adapted levels.}
\vspace{-1em}
\label{fig:SGPR_CIFAR10}
\end{figure}

To understand why, we designed a simple experiment.
We first pretrain ResNet-18-based \citep{he2016deep} networks with different levels $\nu$ of invariance transformations (see the Appendix for a definition of $\nu$).
We then train sparse GP regression \citep[SGPR;][]{pmlr-v5-titsias09a} models on an augmented training set created by propagating ten points sampled from the augmentation distribution through these neural networks.
The samples are generated at different levels of invariance $\nu$, not necessarily matching the levels of the pretrained NNs.
We plot the results in Fig.~\ref{fig:SGPR_CIFAR10}: when the network is trained at a small invariance level $\epsilon$, the performance of the SGPR model is highest at an invariance level of $0.01$, and rapidly drops off for larger invariances (note the logarithmic x scale).
We see a similar result for the network trained at a level of $0.1$.
Finally, when the network is trained at the same level that the orbit points for the SGPR model are sampled at (`adapted'), we see that added invariance helps the accuracy, with no steep drop off in accuracy for larger invariances.
Therefore, adding invariance does help, but only when the network has already been adapted to that invariance. Currently, in our method this coadaptation is prevented by the current need for coordinate ascent training (Sec. \ref{sec:training_scheme}).

This experiment indicates that for datasets requiring larger neural networks, we are in a difficult position.
We need to adapt the feature extractor jointly with the invariances.
However, this approach leads to pathologies as the neural network parameters are not protected from overfitting (\cite{ober2021promises}, see Sec. \ref{sec:training_scheme}), which we previously mitigated with coordinate ascent.
Therefore, relying on the marginal likelihood to learn invariances with a large feature extractor can easily lead to unwanted behavior -- this behavior prevents us from learning these invariances as easily as the marginal likelihood promises.
We believe that ongoing research in Bayesian deep learning will alleviate this problem. Bayesian neural networks with methods for marginalising over lower layers too, thus protecting them against overfitting, will render our approach more easily applicable. Such advances will allow us to learn invariances more easily and on more complex tasks than we did for the MNIST and PCAM datasets.

\section{CONCLUSION}
Neural networks depend on good inductive biases in order to generalise well. Practitioners usually -- successfully -- handcraft inductive biases, but the idea of learning them from data is appealing. Might we automate the modelling pipeline, moving from hand-crafted models to data driven models; much like we replaced hand-crafted features with learned features in deep neural networks? This work proposes one step in this direction. Inspired by Bayesian model selection we employ the marginal likelihood for learning inductive biases. We avoid the intractability of the marginal likelihood for neural networks by using Deep Kernel Learning. This enables us to leverage previous work on invariance learning in GPs for learning data augmentation in neural networks. We learn useful invariances and improve performance, but encounter challenges when optimising our models. We introduce a new sampling-based bound to the ELBO allowing for inference for the Softmax likelihood, the natural choice for classification tasks, hereby alleviating some of the optimisation difficulties. Others we identify as fundamental limitations of the Bayesian last layer approach.

\textbf{Societal Impact: } 
This work is situated within basic research in probabilistic ML and, as such, bears all the risks of automation itself: harmful redistribution of wealth to those with access to compute resources and data, loss of jobs, and the environmental impact of such technologies. In fact, our model is more computationally heavy than a standard neural network with hand-tuned data augmentation. However, in the long term, automatic model selection has the potential to \textit{reduce} the need for hyperparameter tuning, which usually dramatically exceeds the resources needed for training the final model.   

\subsubsection*{Acknowledgements}
MJ is supported by a research grant from the Carlsberg Foundation (CF20-0370). SWO acknowledges the support of the Gates Cambridge Trust for his doctoral studies.

{
\bibliographystyle{abbrvnat}
\bibliography{bib-file}
}

\newpage
\appendix
\onecolumn
\makesupplementtitle
\section{IS THE MARGINAL LIKELIHOOD NECESSARY?}
\begin{figure}[h!]
    \centering
   \includegraphics[width=0.49\hsize]{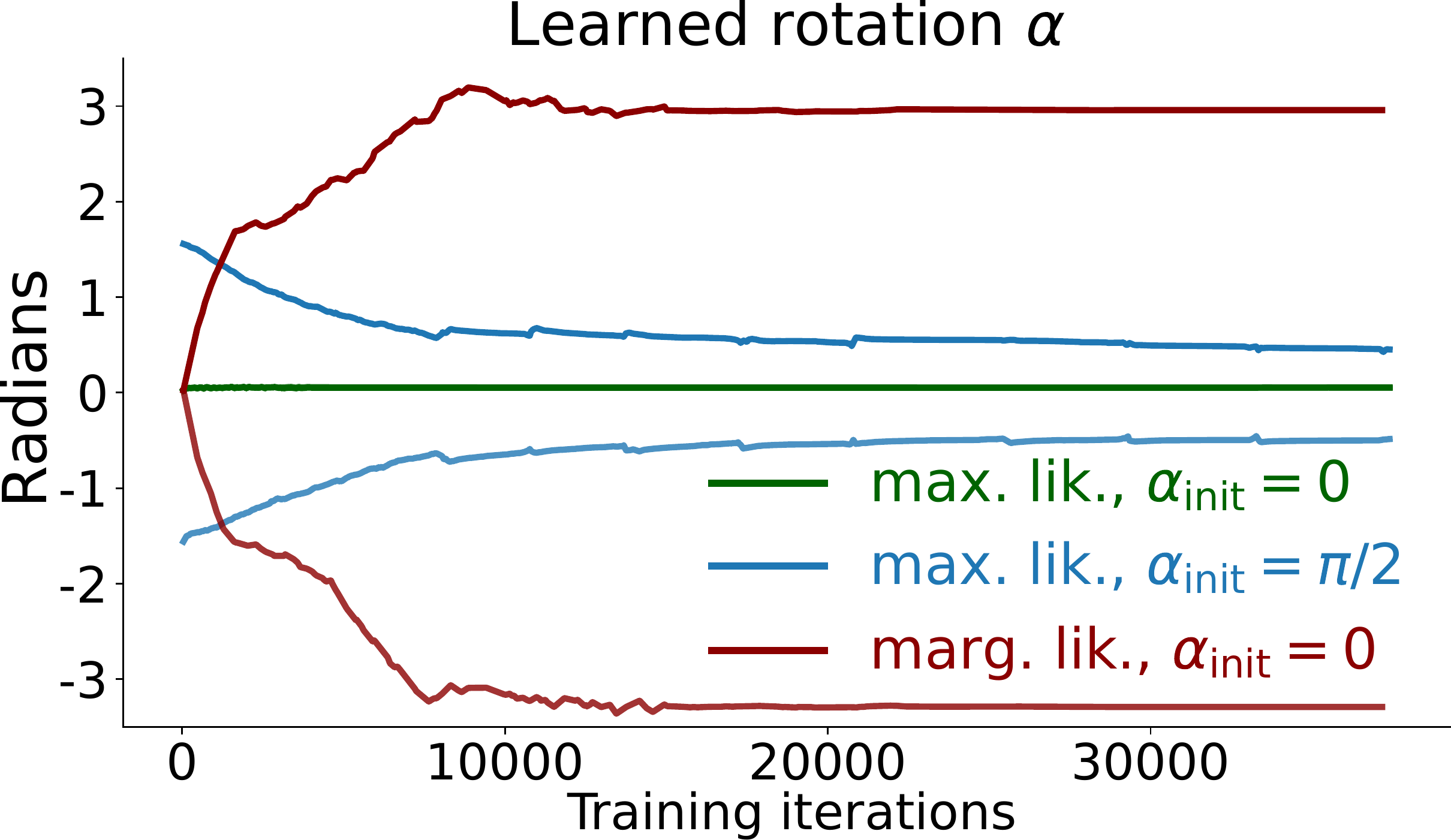}  
  \caption{Max. likelihood (green, blue, \textit{collapsing}) and marg. likelihood (red, \textit{useful}) invariances. }
 \label{fig:MLE_DA_collapse}
 \end{figure}

\thispagestyle{empty}
Sec.~\ref{sec:introduction} motivated the marginal likelihood for invariance learning. Given that this loss function is notoriously difficult to evaluate, we verify experimentally that using it is indeed \emph{necessary}, i.e. that the standard maximum likelihood loss is insufficient. Fig.~\ref{fig:MLE_DA_collapse} shows invariances learned on rotated MNIST (rotMNIST, see Sec.~\ref{sec:rotMNIST} for a description of the dataset) by using a neural network with maximum likelihood loss for two initialisations (blue, green). They collapse as suggested by the theory. The marginal likelihood solution (red) instead identifies appropriate invariances. 

\section{EXPERIMENTAL DETAILS}
Exploiting the ideas from Sec.~\ref{sec:training_scheme}, we start by training convolutional neural networks (CNNs, see below for architecture details). After pre-training the CNN, we replace the last fully connected layer with a GP and continue training. In the non-invariant case we train all parameters jointly from here. When learning invariances, we iterate between updating the GP variational- and hyperparameters, and the neural network weights.

\subsection{MNIST variations}
We here summarise the training setups for the experiments on MNIST variations, i.e. MNIST subsets (Sec.~ \ref{sec:MNIST}) and rotated MNIST (Sec.~ \ref{sec:rotMNIST}). We start by outlining the shared neural network architecture and will then list the hyperparameter settings for MNIST and rotMNIST, respectively.  

\paragraph{The CNN architecture} used in the (rot)MNIST experiments is depicted in Table \ref{table:CNN_arch}. For rotated MNIST, we train the model for $200$ epochs with the Adam optimiser (default parameters). For the MNIST subsets, we train for $60$k iterations which corresponds to $200$ epochs for the full dataset and respectively more epochs for smaller subsets. The remaining parameters are the same in all experiments: batch size $200$, learning rate $0.001$, no weight decay, other regularisation or data augmentation. In the pre-training phase we minimise negative log-likelihood, for updates during coordinate ascent we use the ELBO as a loss-function.

\begin{table*}[h]
\centering
\begin{tabular}{|l|l|}
\hline
\textbf{Layer}   & \textbf{Specifications} \\
\hline
Convolution & filters=20, kernel size=(5, 5),  padding=same, activation=ReLU \\
Max pooling & pool size=(2, 2), stride=2\\
Convolution & filters=50, kernel size=(5, 5),  padding=same, activation=ReLU \\
Max pooling & pool size=(2, 2), stride=2 \\
Fully connected & neurons=500, activation=ReLU \\
Fully connected & neurons=50, activation=ReLU \\
\hdashline
Fully connected & neurons=10, activation=Softmax\\
\hline
\end{tabular}
 \caption{Neural network architecture for MNIST variations. After pre-training, the last fully connected layer (below dashed line) is replaced with a GP layer for the deep kernel models.} \label{table:CNN_arch}
 \end{table*}
 
  \paragraph{Hyperparameter initialisation for the MNIST subset experiments}
are lengthscale	$10$, likelihood variance $0.05$, kernel variance $1$ (fixed likelihood and kernel variance for the invariant model, see Sec.~\ref{sec:MNIST}), posterior variance $0.01$. We use $1200$ inducing points which we initialise by first passing the images through the neural network, then using the `greedy variance' method \citep{burt2020gpviconv} on the extracted features. For the smallest dataset MNIST312 we use $312$ inducing points only. The batch size is $200$ and we choose learning rate $0.001$ for the Adam optimiser. For the invariant models, the orbit size is $120$ and affine parameters are initialised at $\phi_{min} = \phi_{max}=0.02$, i.e. we initialise with a small invariance. Without this initialisation we encountered occasional numerical instabilities (Cholesky errors) on the small dataset runs. During coordinate ascent (InvDKGP models) we toggle between training GP and CNN after $25$k steps.
 
 \paragraph{Hyperparameter initialisation for the rotMNIST experiment} as follows: For all models, we initialise kernel variance $1$ (fixed at $1$ for $M9$, see Sec.~\ref{sec:MNIST}) and posterior variance $0.01$. We use $1200$ inducing points. For the invariant models, the orbit size is $120$ and affine parameters are initialised at $\phi_{min} = \phi_{max}=0$, i.e. invariances are learned from scratch. When using coordinate ascent (InvDKGP models, M8 \& M9) we toggle between training GP and CNN after $30$k steps. We train different models for a different number of iterations, all until the ELBO has roughly converged. Batch size $200$ is used for all models. The remaining initialisations differ between models and are summarised in Table \ref{table:rotMNIST_inits}.
 																		
\begin{table*}[h]
\centering
\begin{tabular}{|l|l|l|l|l|}  
    \hline
     & \textbf{Model}       & \textbf{Lengthsc.}    &\textbf{Lik. var.} & \textbf{LR (decay)} \\
    \hline 
    M2 & Non-inv. Shallow GP + Gaussian &   10  & 0.02 & 0.001  \\
    M3 & Non-inv. Shallow. GP + Softmax &   10   &  -  & 0.001  \\
    M4 & Inv. Shallow GP + Gaussian &   10        & 0.05 & 0.001  \\
    M5 & Inv. Shallow. GP + Softmax &     10       &  -  & 0.001  \\
    \hline 
    M6 & Non-inv. Deep Kernel GP + Gaussian &   10    & 0.05  & 0.001 \\
    M7 & Non-inv. Deep Kernel GP + Softmax &    20    &  -  & 0.001 \\
    M8 & Inv. Deep Kernel GP + Gaussian &   50        &  0.05 (F) & 0.003 (steps / cyclic) \\
    M9 & Inv. Deep Kernel GP + Softmax &      9      &  - & 0.003 / 0.0003 (s / c)  \\
    \hline 
\end{tabular}
\caption{Training settings for rotMNIST models: Kernel lengthscale and likelihood variance initialisations ('F' indicates a fixed likelihood variance, see Sec.~\ref{sec:MNIST}). The learning rate column ('LR') also indicates whether the learning rate was decayed in the GP/CNN update phases of coordinate ascent. For the `steps'(s)  decay, we divide by $10$ after $50$\% and again $75$\% of iterations, for the `cyclic'(c) decay, learning rates are: [LR/$100$, LR/$10$, LR, LR/$10$, LR/$100$]. These training hyperparameters are determined using a validation set (see Sec.~\ref{sec:rotMNIST}).} \label{table:rotMNIST_inits}
\end{table*}

 \subsection{PCam}
 \paragraph{The CNN architecture}
 is a VGG-like convolutional neural network\footnote{We closely follow \url{https://geertlitjens.nl/post/getting-started-with-camelyon/}.} described in Table \ref{table:pcam_arch}. The model is trained for $5$ epochs using the Adam optimiser with batch size $64$. We use learning rate $0.001$ which we divide by $10$ after $50$\% and again $75$\% of training iterations. In the fully connected block we use dropout with $50$\% probability when pre-training.  Dropout is disabled when training the deep kernel models. 
 
\begin{table*}[h]
\centering
\begin{tabular}{|l|l|}
\hline
\textbf{Layer}   & \textbf{Specifications} \\
\hline
 Convolution  & filters=16, kernel size=(3, 3), padding=valid, activation=ReLU \\
 Convolution & filters=16, kernel size=(3, 3), padding=valid, activation=ReLU \\
Max Pooling & pool size=(2, 2), strides=2 \\
Convolution & filters=32, kernel size=(3, 3), padding=valid, activation=ReLU \\
Convolution  & filters=32, kernel size=(3, 3), padding=valid, activation=ReLU \\
Max Pooling  & pool size=(2, 2), stride=2 \\
Convolution & filters=64, kernel size=(3, 3), padding=valid, activation=ReLU \\
Convolution & filters=64, kernel size=(3, 3), padding=valid, activation=ReLU \\
Max Pooling &  pool size=(2, 2), stride=2 \\
Fully Connected  & neurons=256, activation=ReLU \\
Dropout  &  probability=0.5 \\
Fully Connected   &  neurons=50, activation=None \\
Dropout  & probability=0.5 \\ 
\hdashline
Fully connected & neurons=2, activation=Softmax\\
\hline
\end{tabular}
 \caption{Neural network architecture for PCAM. After pre-training, the last fully connected layer (below dashed line) is replaced with a GP layer for the deep kernel models and dropout is disabled.} \label{table:pcam_arch}
 \end{table*}

\paragraph{Hyperparameters} for the deep kernel GP experiments on PCam are: lengthscale $10$ ($1$ for the learned invariance model), kernel variance $1$, posterior variance $0.01$. We use $750$ inducing points which we initialise as in the previous experiments. The batch size is $32$. For PCAm we use coordinate ascent for all models since this improves training stability. Learning rates are $ 0.001$ for the GP update steps and $0.0001$ for the CNN updates, no LR decay. We toggle between the two coordinate ascent phases after $50$k and $75$k iterations in the non-invariant and invariant case, respectively. For the invariant models the orbit size is $20$ and we initialise the rotation invariance with 
$\phi_{min/max} = \alpha_{min/max}= \pm \pi/10$.

\subsection{CIFAR-10}
Throughout our experiments, we train on a subset of 45,000 points from the full CIFAR-10 \citep{krizhevsky2009learning} training set and report results on the remaining 5,000 points, as a validation set.

\paragraph{The model} we use is a sparse GP regression (SGPR; \citet{pmlr-v5-titsias09a}) model with a sum kernel corresponding to a Monte Carlo estimate of the kernel of Eq.~\ref{eq:invariant_kernel}, using an automatic relevance determination (ARD) squared exponential kernel as a base kernel. 
We achieve this by sampling 10 points from the full orbit for each data point, and propagating the points through the pretrained feature extractor. 
For the feature extractor, we choose a ReLU ResNet-18 architecture \citep{he2016deep} with an output dimension of 50, using the post-ReLU features.
Therefore, for our training set, we end up with a set of $45,000 \times 10 \times 50$ datapoints, where we sum over the 10 orbit samples.

\paragraph{Hyperparameters} were chosen as follows.
We pretrain the ResNet-18 by adding an additional fully-connected layer with softmax activations.
We train for 160 epochs with a batch size of 100 and the Adam optimizer \citep{kingma2015adam}, starting with a learning rate of 0.001, which we step down by a factor of 10 at epochs 80 and 120. 
We train the network without weight decay.
The SGPR model was subsequently trained for a maximum of 1000 steps, using the Scipy optimizer provided in GPflow \citep{GPflow2017}.
During training, we initially set the jitter to 1e-6, which we increased by a factor of ten if the Cholesky decomposition failed.
For the SGPR model, we use 1000 inducing points, initialised as above.
We found empirically that the likelihood variance did not have a significant impact on the results; we therefore fixed it to 0.01.
Recalling that $\phi = (\alpha, s^x, s^y, p^x, p^y, t^x, t^y)$, we parameterize the transformation by considering the ``transformation level'' $\nu$ such that 
\begin{align}
    \phi_\text{max} &= (0, 1, 1, 0, 0, 0, 0) + \nu\times(\pi, 1, 1, 1, 1, 1, 1), \\
    \phi_\text{min} &= (0, 1, 1, 0, 0, 0, 0) - \nu\times(\pi, 1, 1, 1, 1, 1, 1).
\end{align}
For the ``$\epsilon$'' setting of the transformation level, we assign $\nu=0.01$.
We chose a non-zero value to ensure that any reduction in performance would be due to a different value of $\nu$, and not because of the lack of presence of the image interpolator in the pretraining (see Sec.~\ref{sec:PCAM}).


\end{document}